\documentclass[10pt,journal]{IEEEtran}
\pdfoutput=1
\makeatletter
\DeclareRobustCommand\onedot{\futurelet\@let@token\@onedot}
\def\@onedot{\ifx\@let@token.\else.\null\fi\xspace}
\def\eg{\emph{e.g}.} 
\def\ie{\emph{i.e}.}

\def\etal{\emph{et al}.}
\makeatother

\usepackage{pifont}
\usepackage{gensymb}
\usepackage{amsfonts}
\usepackage{amsmath}
\usepackage{amssymb}
\usepackage{subfigure}
\usepackage{multirow}
\usepackage{color}
\usepackage{cite}
\usepackage{url}
\usepackage{hyperref}
\hypersetup{
    colorlinks=true,
    linkcolor=blue,
    filecolor=magenta,      
    urlcolor=cyan,
}

\ifCLASSOPTIONcompsoc
  \usepackage[nocompress]{cite}
\else
  \usepackage{cite}
\fi

\ifCLASSINFOpdf
  \usepackage[pdftex]{graphicx}
  \graphicspath{{./img/}}
\else
\fi
\begin{document}
\title{DAVE: Deep Audio-Visual Embedding \\for Dynamic Saliency Prediction}

\author{Hamed~R.Tavakoli,~\IEEEmembership{Member,~IEEE,}
        Ali~Borji,
        Esa Rahtu,
        and~Juho~Kannala

\IEEEcompsocitemizethanks{\IEEEcompsocthanksitem H. R. Tavakoli was with the Department
of Computer Science, Aalto University, Espoo, Finland. He is currently a principal scientist at the Nokia Technologies, Finland. (e-mail: hamed.rtavakoli@gmail.com)
\IEEEcompsocthanksitem Ali Borji is a scienior scientist.
\IEEEcompsocthanksitem Esa Rahtu is with Tampere University, Tampere, Finland.
\IEEEcompsocthanksitem Juho Kannala is with the Department
of Computer Science, Aalto University, Espoo, Finland.}
\thanks{Manuscript received April 19, 2005; revised August 26, 2015.}}

%
%

\markboth{Journal of \LaTeX\ Class Files,~Vol.~14, No.~8, August~2015}%
{Shell \MakeLowercase{\textit{et al.}}: Bare Advanced Demo of IEEEtran.cls for IEEE Computer Society Journals}

\IEEEtitleabstractindextext{%
\begin{abstract}

This paper studies audio-visual deep saliency prediction. It introduces a conceptually simple and effective \textit{D}eep \textit{A}udio-\textit{V}isual \textit{E}mbedding for dynamic saliency prediction dubbed ``DAVE" in conjunction with our efforts towards building an Audio-Visual Eye-tracking corpus named ``AVE". 
Despite existing a strong relation between auditory and visual cues for guiding gaze during perception, video saliency models only consider visual cues and neglect the auditory information that is ubiquitous in dynamic scenes. Here, we investigate the applicability of audio cues in conjunction with visual ones in predicting saliency maps using deep neural networks. To this end, the proposed model is intentionally designed to be simple. Two baseline models are developed on the same architecture 
which consists of an encoder-decoder. The encoder projects the input into a feature space followed by a decoder that infers saliency.
We conduct an extensive analysis on different modalities and various aspects of multi-model dynamic saliency prediction. 
Our results suggest that (1) audio is a strong contributing cue for saliency prediction, (2) salient visible sound-source is the natural cause of the superiority of our Audio-Visual model, (3) richer feature representations for the input space leads to more powerful predictions even in absence of more sophisticated saliency decoders, and (4) Audio-Visual model improves over 53.54\% of the frames predicted by the best Visual model (our baseline). Our endeavour demonstrates that audio is an important cue that boosts dynamic video saliency prediction and helps models to approach human performance. The code is available at \href{https://github.com/hrtavakoli/DAVE}{https://github.com/hrtavakoli/DAVE}

\end{abstract}

\begin{IEEEkeywords}
Multimodal video saliency, dynamic visual attention, audio-visual saliency, deep learning
\end{IEEEkeywords}}

\maketitle

\IEEEdisplaynontitleabstractindextext
\IEEEpeerreviewmaketitle

\ifCLASSOPTIONcompsoc
\IEEEraisesectionheading{\section{Introduction}\label{sec:introduction}}
\else
\section{Introduction}
\label{sec:introduction}
\fi

\IEEEPARstart{C}{omputational} Models of Visual Attention (CMVA) attempt to explain the observed behavior of primate visual attention and make predicions. Saliency models are a sub-category of such computational models~\cite{John2011}. They are used to predict the most prominent area of images (static/dynamic) in terms of
density maps (aka saliency maps) obtained from human eye fixations during task-free scene viewing. Such models have been extensively studied within various disciplines, including, psychology, cognitive vision sciences, and computer vision~\cite{Itti2000}.

In computer vision, while a plethora of models have been reported for image saliency, fewer works have been concerned about video saliency (aka dynamic saliency). Video saliency models can be utilized in various applications such as video analysis and summarization~\cite{Marat2007}, stream compression~\cite{Hadi2014}, augmented and virtual reality~\cite{Sandor2010,ZhangSaliencyDI}, etc.
Many of such applications are naturally multi-modal,
containing a dynamic sequence of images and audio. 
In spite of the existing evidence on the correlation between auditory
and visual cues and their joint contribution to attention~\cite{Burg2008},
to date, most of the video saliency models neglect audio and heavily rely on spatio-temporal visual cues as the only source of information.

In this paper, to the remedy the above shortcoming, we focus on multi-modal (audio-visual) modeling of saliency over videos with the emphasis on end-to-end trainable deep saliency models, which have been dominating the field recently~\cite{Borji:survdeep:2018}. For this purpose, we (1) construct a relatively large database by pooling several databases together, (2) propose a multi-modal saliency model, and (3) analyze and discuss the contribution of audio in saliency prediction.

We construct a database by combining data from various sources to assist the training of deep audio-visual models. In our database, the videos are accompanied by eye tracking data gathered under free-viewing condition when both audio and video have been presented to the viewers. We categorize the videos into three types. This further enables us to perform a fine-grained analysis.

The proposed deep multi-modal saliency model uses both audio and visual information to predict salience. The model is purposely designed to be simple to facilitate the analysis of the contribution of each modality. Our model is also consistent with the famous feature integration theory~\cite{TREISMAN1980}, which has been the basis of many saliency models so far. To our knowledge, we are one of the few who proposes an end-to-end trainable deep model for dynamic audio-visual saliency.

Along with the multi-modal model, we build two baselines including an audio saliency model and a visual saliency model. The baseline models share the same architecture with our multi-modal model, except that they rely only either on video or audio. They share the same training procedure to facilitate the analysis of the contribution of each information cue.

We perform a series of extensive analysis of the model and modalities (audio, visual, audio-visual) over each video category and overall. The audio-visual model outperforms our both baselines over all scores; overall and over individual categories. Contextual analysis of the model performance over the location of sound source reveals that the audio-visual model behaves similar to humans in attending to the location of sound source. Our analysis demonstrates that audio is a significant contributing cue inline with behavioural studies~\eg~\cite{Burg2008}.

We also conduct a comparison of existing video saliency models. The proposed model outperforms all of the video-only models. This further demonstrates that audio is an important signal that can boost video saliency prediction and help getting closer to human performance. 

In a nutshell, our main contributions include: 
\begin{itemize}
    \item Constructing a database for audio-visual saliency prediction,
    \item Providing video categorical annotation for the data to enhance model analysis with respect to stimulus type,
    \item Introducing a deep audio-visual saliency model for video, 
    \item Assessing the contribution of each modality (audio, video, and audio-video) using deep saliency models, 
    \item Comparison and analysis of the proposed audio-visual saliency model against the existing video saliency models.
\end{itemize}

\section{Related Works}

Humans are intelligent multi-sensory creatures, capable of spotting and focusing on a certain parts of audio or visual stimului in a cluttered environment (\ie, have attentional behavior). It is, hence, unsurprising that inspired by such observations
psychologists and neuroscientists have been studying attentional mechanisms underlying auditory, visual, and auditory-visual attention.  
The recorded seminal ideas and works on attention can be traced back to the 19th century~\cite{James1890}.
Several decades of research on attention mechanisms have amassed a rich literature on the topic. Covering the whole literature is, thus, beyond the scope of this paper. Instead, to stress the need for multi-modal attention models, the following subsections provide a brief account of relevant studies.
We refer the reader to~\cite{Carrasco2011,John2011,Borji2013,Kaya2016,Borji:survdeep:2018} for further information on these topics. 

\subsection{Saliency Prediction vs. Saliency Detection}

We review and discuss saliency prediction by computational models of saliency versus detection and segmentation of a salient object (salincy detection) for the sake of clarity. The two tasks are well-established, known to be relevant and yet distinct. This relation and distinction is discussed in depth in~\cite{Li2014} for image saliency and can be extended to video saliency following the same principle. 

From a theoretical perspective, saliency prediction with computational models of saliency correspond to the mechanism that results in deployment of human attention to a region in an image or video. On the other hand, the saliency detection or segmenting the salient object is more relevant to the pereptual grouping~\cite{Buhmann1999} as part of a recognition pipeline where visual factors can be used to group parts of an image that most likely belong to a single object. 

From a computer vision point of view, the two tasks are distinguished by their ground-truth data. For saliency prediction, the ground-truth data is a fixation density map (saliency map) obtained from eye movements of human observers. On the other hand, for saliency detection, the ground truth is an object mask that identifies the most salient object.

The two tasks are, however, tightly linked as demonstrated in~\cite{borji2013stands,Li2014}. In such a scenario, a saliency map is computed to infer a salient seed point. Then, an object segment can be inferred in a second step using the seed point, saliency map, and the visual features. Examples of such pipelines can be found in the literature \eg~\cite{Fu2008,Mishra2009,Kruthiventi2016}. There has been, however, a trend towards completing the visual grouping pipeline by unifying saliency prediction, segmentation of object instances and recognition tasks~\eg~\cite{TavakoliL17}. Similar efforts for detecting and ranking multiple salient objects have been pursued in the saliency detection community~\eg~\cite{Islam_2018_CVPR}. We refer the readers to~\cite{Fan2019VideoSal} in regards to state-of-the-art in salient object detection and the potential bridge between saliency detection and saliency prediction as they provide excellent object annotation for a subset of the video saliency prediction database of~\cite{wang2018cvpr}.
The data, however, has been present to observers with no audio during eye-tracking task. Hence, it is not useful for multi-modal saliency studies.

To our knowledge, so far there has not been efforts in the computer vision saliency detection community for multi-modal salient object segmentation. Nonetheless, our focus area is computational models of saliency prediction.

\subsection{Interplay Between Visual and Auditory Attention}

Auditory attention regards mechanisms of attention
in the auditory system. A good example of such mechanisms is
the famous cocktail party problem~\cite{Cherry1953}. 
The behavioral studies on the formation of 
auditory attention have been interested in the subject's response 
to an auditory stimulus, \eg~\cite{Maccoby1966,Bartgis2003}.
From neurophysiological point of view, mechanisms
of auditory attention and its influence on auditory cortical 
representations have been of interest, \eg~\cite{Mesgarani2012}. 

Parallel to studies on auditory attention, numerous works have 
explored visual attention. The span of behavioral studies on visual attention is broad and covers a wide range of experiments on
primates and humans. 
In such experiments, an observer is often presented with
a stimulus and his neural and/or behavioral responses are recorded (\eg using single unit recording, brain imaging, or eye tracking)~\cite{Carrasco2011}.

With respect to multimodal attention, some behavioral studies have investigated the role of audio-visual components during sensory development in infants~\cite{Lewkowicz1988}. Richards~\cite{Richards2000} studied attention engagement to compound auditory-visual stimuli and showed an increase in sustained attention development with age progression in infants. Burg~\etal~\cite{Burg2008} conducted experiments to understand the effect of audio-visual events on guiding attention and concluded that audio-visual synchrony guides attention in an exogenous manner in adults.

\subsection{Computational Models of Visual Saliency}

The proposed model follows the recent wave of data-driven saliency modeling based on deep neural networks, \eg~\cite{Huang2015ICCV,Kummerer2014b,Cornia2016}. Such data-driven approaches fall within the broader 
category of models based on the feature integration theory (FIT)~\cite{TREISMAN1980}. That is the visual
input is mapped into a feature space and a saliency map is obtained by combining the features. 
Many of the saliency models can be categorised within this broad category, though they may follow different
computational schemes~\cite{Hamed2014phd}. The deep neural saliency models can be seen as an encoder-decoder architecture.
First, a feed-forward convolutional neural network (CNN) projects the input to a feature space (encoder). Then, a second neural architecture combines the features to form a saliency map (decoder). The decoder consists of series of convolution operations or more complicated recurrent networks (\eg~\cite{cornia2018,wang2018cvpr}). Our approach follows the same principle, however, it is applied to both audio and video modalities.

\subsection{Computational Models of Auditory Saliency}

The computational modeling of auditory saliency is
a relatively younger field than visual salience modeling. 
Some works have been inspired by visual saliency techniques. For example,~\cite{Kayser2005} adopts the model of~\cite{Itti2000} to audio domain and produces auditory saliency maps. 
An auditory salience map identifies the presence of a salient sound over time. Some models are, however, rooted in the biology of the auditory system or extend saliency map idea for audio-based tasks.

Wrigley~\etal~\cite{Wrigley2004} proposed a model in which a network of neural oscillators performs auditory stream segregation on the
basis of oscillatory correlation. To be concise, 
the model processes the audio input to simulate auditory 
nerve encoding. Periodicity information is extracted using
a correlogram and noise segments are identified. This information
is passed through an array of neural oscillators in which each
oscillator corresponds to a particular frequency channel. The
output from each oscillator is connected via excitatory links to
a leaky integrator that decides which oscillator and in consequence 
input channel should be attended. 

Oldoni \etal~\cite{Oldoni2013} proposed an auditory attention model for designing better
urban soundscapes. Their model is largely inspired
by the FIT~\cite{TREISMAN1980}. The audio signal is converted
to a 1/3-octave band spectrogram. Intensity, spectral contrast, and
temporal contrast are computed from the spectrogram. This process
corresponds to the peripheral auditory processing and results in a feature vector that forms a saliency map,
which will be used to identify the sounds that will be heard.

While having some commonalities with
vision based models of attention in the first layer, this model goes beyond saliency map approaches and extends such models to task-based top-down driven attention.

The audio branch in the proposed framework follows the principle of FIT for the sake of consistency with the visual part. The main
difference with the auditory models of saliency is the 
application of deep neural networks.

\subsection{Models of Audio-Visual Saliency}

To the best of our knowledge, only few multi-modal saliency works exist. Some of the early works include ~\cite{CoutrotModel2014,CoutrotModel2015,CoutrotModel2016}, which are extended by~\cite{multiECCV2018}. 
In~\cite{CoutrotModel2014}, static and dynamic low-level video features are extracted using Gabor filters. The faces are also segmented using a semi-automatic segmentation tool~\cite{Bertolino2012} interactively as part of the visual feature space. For the audio track of video frames a speaker diarization technique is proposed based on voice activity detection, audio speaker clustering, and motion detection. This information is then combined with visual information to obtain a saliency map. In~\cite{CoutrotModel2015,CoutrotModel2016}, this framework is improved by adding annotation of body parts. The need for manual face and body part segmentation limits the applicability of these models in real-world scenarios.

Boccignone~\etal~\cite{multiECCV2018} follow a similar path for automating saliency prediction in social interaction scenarios. They extract several priority maps, including (1) spatio-temporal saliency features using~\cite{Seo2009}, (2) face maps from a convolutional neural network (CNN) face detector, and (3) the active speaker map from automatic lip sync algorithm of~\cite{Chung16a}. Once the maps are extracted, a sampling scheme is employed to find attention attractors. 
Instead of relying on a complicated sampling scheme and multiple feature maps, we directly learn the mapping using a deep neural network. 

Our model is distinct from aforementioned audio-visual saliency models because (1) contrary to existing models that only focus on conversations and faces, our model is applicable to any scene type, and (2) it is a single end-to-end trainable framework for the multi-modal saliency prediction.

\section{Audio-Visual Eye-tracking}

Audio-visual eye tracking data is required in order to build a multi-modal saliency model, \eg~for training and evaluation, and investigating the contribution of modalities. In other words, one has to record fixations of observers while both audio and visual modalities are presented simultaneously and synchronously.

Unfortunately, the field lacks a large corpus of fixation data where the audio and video stimuli have been presented simultaneously. The current large scale video saliency databases focus only on visual cue during the eye tracking experiments and do not present the audio to observers during the data gathering procedure. Table~\ref{tab:databases} describes some of the existing databases with emphasis on the presence and the quality of auditory cue during eye tracking experiment. As shown, only a limited number of video sequences with fixations is available that meets the required criteria (i.e., presence of video and audio during eye tracking). To remedy this, we pool the existing data from~\cite{Mital2011DIEM},~\cite{Coutrot2011erbsound} and~\cite{CoutrotModel2015} together to construct a dataset for model training. We refer to this data as Audio-Visual Eye-tracking (AVE) corpus.

\begin{table*}[!t]
\renewcommand{\arraystretch}{1.3}
\caption{The video databases, the presence of audio during eye tracking experiments and the average number of observers per-sequence is reported.}
\label{tab:databases}
\centering
 \small
\begin{tabular}{r|c|c|c|p{8cm}}
\hline
 \textbf{Dataset} &  \textbf{\# Videos} &  \textbf{Audio Pres.} & \# Observers & \hfil  \textbf{Notes}  \\
\hline
CRCNS~\cite{Itti2004} & 50 & \ding{55} & 15 & Fixations of observers who were instructed to follow the most important objects.\\
AIE~\cite{MatheSminchisescuPAMI2015} & 1707 & \ding{55} & 16 & The largest eye tracking corpus for action recognition, in which eye-tracking data is gathered under task-based and task-free conditions.\\

DHF1K~\cite{wang2018cvpr} & 1000 & \ding{55} & 17 & A large scale free-viewing database with various video categories.  \\
DAVSOD~\cite{Fan2019VideoSal} & 266 & \ding{55} & 17 & Subset of DHF1K anotated for saliency segmentation. \\
LEDOV~\cite{Jiang2018ECCV} & 538 & \ding{55} & 32 & A free-viewing database with various video categories and increased number of observers.  \\
DIEM~\cite{Mital2011DIEM} & 84 & \ding{51} & 50 & An audio-visual eye tracking database with largest number of observers.  \\
Coutrot~1~\cite{Coutrot2011erbsound}  & 60 & \ding{51} & 18 & Experiments with multiple auditory conditions to study the contribution of auditory cue in visual perception.\\
Coutrot~2~\cite{CoutrotModel2015}  & 15 & \ding{51} & 20 & Experiments with and without sound over conversations.\\
AVE (\textbf{ours}) & 150 & \ding{51} & 45 & Subset of the sequences of DIEM, Coutrot~1, and Coutrot~2; with extra scene type categorization and train/val/test splits.\\
\hline
\end{tabular}
\end{table*}

To construct the database, we ensure the auditory cue is of the visual content and the audio signal is synchronized with video. 
From~\cite{Mital2011DIEM}, we only select 76 videos for which the audio cue is provided publicly and we are sure it has been played during eye tracking. We use all the sequences from \cite{Coutrot2011erbsound} and \cite{CoutrotModel2015}. In total, we have 150 videos. The video sequences are on average $62$ seconds long (minimum: $8$, maximum: $210$, median: $45$ seconds). In total, the dataset consists of $267,109$ frames corresponding to $153.12$ minutes (approximately 2 hours and 33 minutes). 

\begin{figure}[t!]
    \centering
    \includegraphics[scale=0.33]{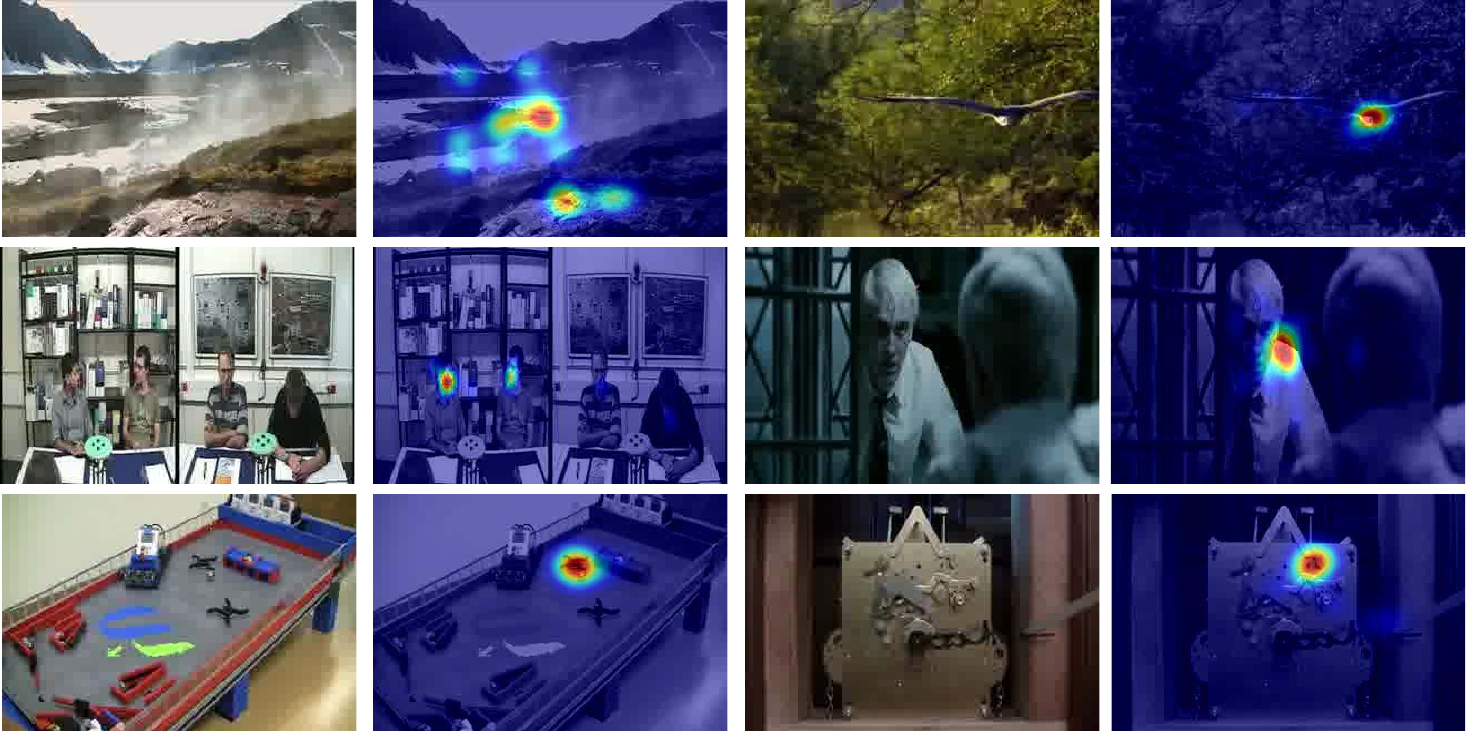}
    \caption{Example frames of each video category with the corresponding ground-truth fixations overlaid. From top to bottom: Nature, Social Events, and Miscellaneous classes. }
    \label{fig:catexamples}
\end{figure}

We categorized the video sequences into three categories based on their content: `Nature', `Social events', and `miscellaneous'. Examples from each video category and fixation density maps are provided in Fig.~\ref{fig:catexamples}. The `Nature' category includes videos of nature and animals in their habitats. `Social events' sequences contain group activities including at least 2 people and covers a wide range of activities such as sports, crowds, conversations, etc. Remaining videos are categorized as `Miscellaneous'. We also divided the videos into 
train/validation/test splits. Table~\ref{tab:nos} summarizes the number of videos in each category per split.

 \begin{table}[t]  
  \renewcommand{\arraystretch}{1.3}
 \caption{Number of sequences in each video category.}
 \label{tab:nos}
 \centering
 \small
 \begin{tabular}{l l l l}
 \hline
 \textbf{Video Type} & \textbf{Train} & \textbf{Valid.}  & \textbf{Test}\\
 \hline
 Nature & 37 & 6 & 10\\
 Social events & 18 & 16 & 11\\
 miscellaneous & 37 & 7 & 7\\
 \hline
 \textbf{Total} (150) & 92 & 29 & 29\\
 \hline
 \end{tabular}
 \end{table}

The average number of subjects per video is 45 (minimum: 18, maximum: 220, median: 31). All fixation data are recorded at 1000 Hz with an SR-EyeLink1000 eye tracker under the free-viewing task. The presentation setup has been slightly different in the above mentioned source datasets. This, however, does not affect us as (1) the gaze data has been appropriately mapped to correct 2D pixel coordinates and (2) correct smoothing according to $1^{\degree}$ viewing angle of each source has been applied.  

\begin{figure}[!t]
    \centering
    \includegraphics[scale=0.65]{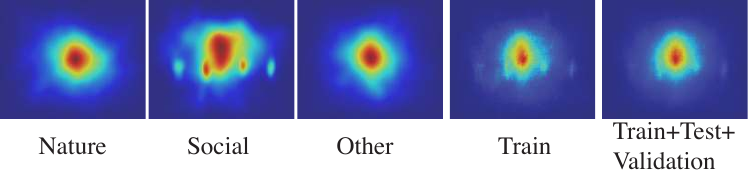}
    \caption{From left to right: Mean Eye Position (MEP) for the different categories in the training set, for the all training sequences (used as a lower bound baseline), and for the entire dataset (train,val,test).}
    \label{fig:cbias}
\end{figure}

We visualized the mean eye position (MEP) map over all frames as well as over different categories in Fig.~\ref{fig:cbias}. As shown in the literature~\cite{Tatler2007}, fixations tend to gravitate more towards the center (a.k.a, center-bias). The mean eye position map (MEP) from the training sequences depicts the center-bias that a model may learn by training on the dataset. It is, hence, a powerful lower-bound baseline.

The upper-bound performance on the database is computed by splitting  subjects into two groups and assessing one group against the other. This is similar to the infinite human analysis in~\cite{Judd2012}, which is a common approach for establishing an upper-bound performance for the saliency models. We call the upper-bound `Human Infinite'.

We summarize the bounds of the database, computed over each category and overall, in Table~\ref{tab:bounds}. The performance bounds will provide us an idea about the performance gap of the models assessed on our data. A model has an acceptable performance if it outperforms MEP. The better a model is, the closer its scores to the human infinite scores are. 

An interesting observation is the higher NSS scores of the human infinite baseline for `social events'. This suggests that there is a higher degree of agreement between observers in terms of where they look. For example, if two people are talking in a video, most of the observers will attend to the same speaker. The MEP has also the lowest prediction score on `social events' indicating that the salient regions are likely off-center and more unique, which is also evident in Fig.~\ref{fig:catexamples}. 

\begin{table}[t]
    \renewcommand{\arraystretch}{1.3}
  \renewcommand{\tabcolsep}{.9mm}  
 \caption{Performance bounds of the dataset: Human Infinite (upper bound) and MEP (lower bound) expected performance over the database.}
 \label{tab:bounds}  
 \footnotesize
 \centering
 \begin{tabular}{c l c c c c c}
 \hline
 \textbf{Cat.}&\textbf{Model Name} & \textbf{NSS} $\uparrow$ & \textbf{AUC Judd} $\uparrow$   & \textbf{sAUC} $\uparrow$   & \textbf{CC} $\uparrow$   & \textbf{SIM} $\uparrow$ \\
 \hline
\multirow{2}{*}{{\textbf{Nature}}} & Human Inf. & \textcolor{red}{3.31} & \textcolor{red}{0.8806} & 
\textcolor{red}{0.7724} & 
\textcolor{red}{0.6961} & 
\textcolor{red}{0.5603} \\
& MEP & \textcolor{blue}{1.76} & 
\textcolor{blue}{0.8696} & 
\textcolor{blue}{0.6864} & 
\textcolor{blue}{0.4714} & 
\textcolor{blue}{0.3682} \\
 \hline
 \multirow{2}{*}{{\textbf{Soc. Ev.}}} & Human Inf. & 
 \textcolor{red}{3.65} & 
 \textcolor{red}{0.8765} & 
 \textcolor{red}{0.7760} & 
 \textcolor{red}{0.6971} & 
 \textcolor{red}{0.5485} \\
& MEP &  \textcolor{blue}{1.35} & 
\textcolor{blue}{0.8196} & \textcolor{blue}{0.6337} & 
\textcolor{blue}{0.3147} & 
\textcolor{blue}{0.2744} \\
 \hline
 \multirow{2}{*}{{\textbf{Misc.}}} & Human Inf. & 
 \textcolor{red}{3.34} & 
 \textcolor{red}{0.8682} & 
 \textcolor{red}{0.7716} & 
 \textcolor{red}{0.6543} & 
 \textcolor{red}{0.5256} \\
& MEP & \textcolor{blue}{1.73} & 
\textcolor{blue}{0.8446} & 
\textcolor{blue}{0.6754} & 
\textcolor{blue}{0.4378} & 
\textcolor{blue}{0.3422} \\
 \hline
 \multirow{2}{*}{{\textbf{Overall}}} &Human Inf. & \textcolor{red}{3.83} & \textcolor{red}{0.8852} &
 \textcolor{red}{0.7877} & 
 \textcolor{red}{0.7140} &
 \textcolor{red}{0.574 } \\
& MEP & \textcolor{blue}{1.59} & 
 \textcolor{blue}{0.8443} & 
 \textcolor{blue}{0.6623} & 
 \textcolor{blue}{0.4029} & 
 \textcolor{blue}{0.3263} \\
 \hline
 \end{tabular}
 \end{table}

\section{Model}

\begin{figure}[th]
    \centering
    \includegraphics[width=8cm,height=3.2cm]{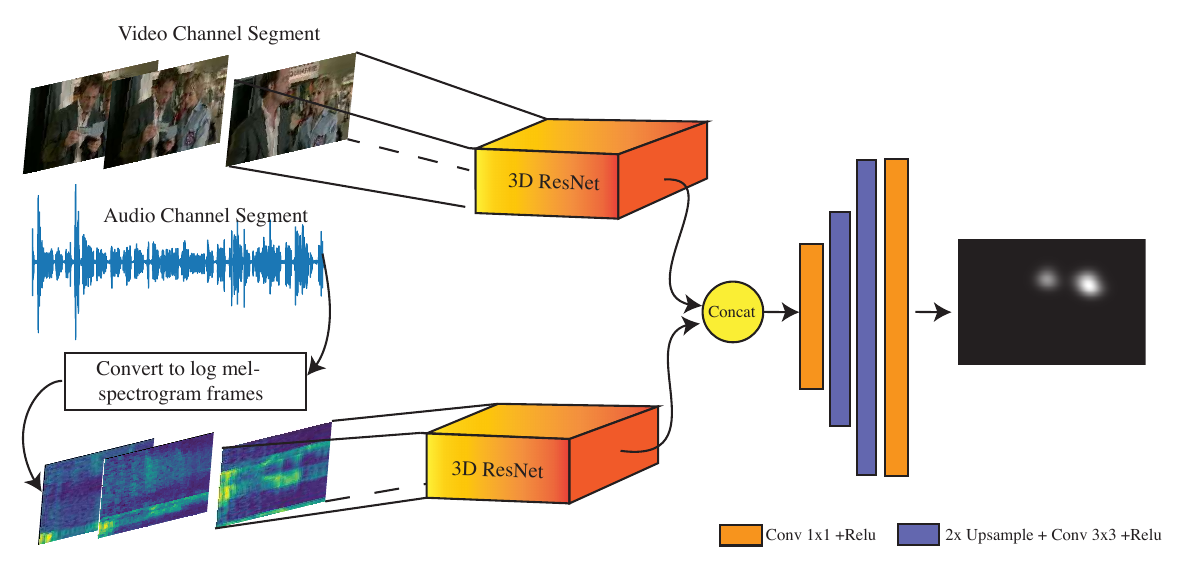}
    \caption{The overall architecture of the proposed Audio-Visual model. The features are prjected into a feature space, concatenated together and a decoder produces the saliency from them.}
    \label{fig:pipeline}
\end{figure}

We propose a model based on the feature integration theory (FIT)~\cite{TREISMAN1980}. That is, the input is projected to a feature space and a saliency map is learnt from this feature space. Our approach follows the learning-based saliency paradigm~\cite{Zhao2011} within the deep learning techniques, which follow encoder and decoder neural architectures.

Fig.~\ref{fig:pipeline} depicts the proposed pipeline for multimodal saliency prediction. We formulate the problem as follows. Given a video segment as a set $\mathrm{V}=\{I, A\}$, consisting of a frame sequence $I$, and an audio signal $A$, the saliency of the video segment can be computed as the following probability distribution:

\begin{equation}
 S = P(S|I, A) = f(I, A),
\end{equation}

\noindent where $f(\cdot,\cdot)$ is a neural network.

As shown in Fig.~\ref{fig:pipeline}, we use a two-stream neural network, one stream for video and another for audio. These streams are based on 3D Convolutional neural networks (3D CNNs), in particular 3D Residual Networks (ResNet)~\cite{Hara2018CVPR}. The video stream processes 16 video frames at a time. The audio stream handles log mel-spectrogram of the audio signal arranged into 16 frames. The obtained representations are concatenated and fed to a series of upsampling and convolution layers to obtain a final saliency map.

To train such a network, we utilize the ground truth saliency distribution obtained from fixation density maps, denoted as $G$, and the saliency distribution predicted by the model, denoted as $S$. We minimize the KL-divergence between the two distributions,
\begin{equation}
    \mathcal{L} (S, G)= \sum_{x\in {X}} G(x)\log(\frac{G(x)}{S(x)}),
\end{equation}

\noindent where $x$ represents the spatial domain of a saliency map.

\subsection{Implementation Details}
In the following, we will explain the technical details of the proposed model and what needs to be taken into account for reproducing the results. Our implementation is available at  \url{https://github.com/hrtavakoli/DAVE}.

\subsubsection{Encoding video using 3D CNNs:} The audio and video backbone of our model is based on the 3D ResNet architecture~\cite{Hara2018CVPR}. We choose the depth to be 18 to keep a balance between audio and video branches as we can not train a very deep network for audio due to dimensions of audio log mel-spectrograms. For the video branch, we initialize the weights from the models pre-trained on the Kinetics dataset~\cite{Kinetics2017} for action classification. The input is of size
$F\times C\times256\times320$, where $F=16$ is the number of frames and $C=3$ is the number of frame channels. The input frames are normalized with the mean and standard deviation of frames from the Kinetics training set.

 \subsubsection{Audio preprocessing:} For the audio signal, we resample the audio to 16KHz and transform it into a log mel-spectrogram with a window length of 0.025 seconds and a hop length of 0.01 seconds with 64 bands. We then convert the transformed audio information into a sequence of successive overlapping frames, resulting in an audio tensor representation of shape $F\times C\times 64 \times 64$, where $C=1$ is the number of channels. This procedure follows the steps of~\cite{Hershey2017}. Since we apply a 3D ResNet adapted from a pre-trained model on images, we replicate the input to have $C=3$ and process $F=16$ audio frames at a time.

\subsubsection{Encoding audio data using 3D CNNs:} To handle audio data, we initialized the  with weights pre-trained on Kinetics dataset~\cite{Kinetics2017}. Then, we re-trained the model for audio classification on the speech command database~\cite{SpeechCmd2018}, where the task is command classification. This dataset consists of $65,000$ one-second long utterances of 30 short words, by several different people. Once the training with this data is done, we use the weights of this network to initialize the audio branch of our saliency model for training.

\subsubsection{Decoding saliency from audio-visual features:} The encoded visual and auditory information using the 3D CNNs are concatenated together. These features are then fed to a 2D convolution layer with kernel size 1, which reduces the feature size by half. This layer is followed by two blocks consisting of a bilinear up-sampling with factor 2, followed by a 2D convolution layer of kernel size 3 and batch normalization. The first block reduces the feature space by half and the second one reduces the feature space by factor of 4. The final layer is a 2D $1\times1$ convolution, resulting in a saliency map of size $32 \times 40$.

\subsection{Training by Dynamic Routing} 

\begin{figure}[t!]
    \centering
    \includegraphics[scale=0.75]{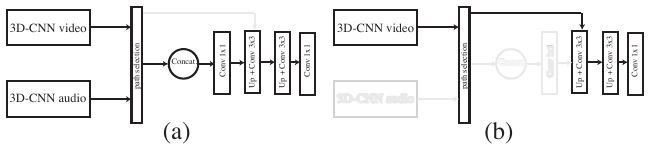}
    \caption{\small{Computational graph during training, (a) handling audio-visual data, (b) handling static images as video frames by replicating them. Path selection decides which part of the graph should be frozen (light gray areas) based on input during training.}}
    \label{fig:graph}
\end{figure}

The lack of stimulus diversity in video databases has been shown to hinder video saliency learning~\cite{wang2018cvpr}. To circumvent this situation, we train our model on data from static saliency datasets as well as video data. To this end, we update the graph dynamically depending on the source of the input. In the case of static images, the image is replicated to form a volume of appropriate number of frames. Then, the audio branch and the section of graph used for mixing the audio and video features are frozen and are ignored in the computational graph during training. In other words, the visual features are directly fed to the first upsampling block, hence just the visual 3D CNN and part of the saliency decoder are updated during the backward pass. A schematic diagram of the computational graph is illustrated in Fig.~\ref{fig:graph} to demonstrate training with static images and audio-visual input.
The model is trained alternatively one epoch for audio-visual input and one epoch for image volumes. 
During training the batch size is 10, learning rate start with $1e^{-3}$ and Adam optimizer is used for a total of 10 epochs. 

\section{Experimental Setup}

We conduct a series of experiments to demonstrate the effectiveness of audio and its contribution by comparing the multimodal model to its corresponding baselines. We, then, compare the proposed model to other video saliency models. For this purpose, we employ the AVE corpus.

\subsection{Evaluation Scores}

Evaluating image saliency models and benchmarks has been studied extensively in the past~\cite{Borji2013ICCV,Judd2012,salMetrics_Bylinskii,TavakoliABL17,Bench2018}. However, there has been less efforts for studying the evaluation of video models. Traditionally, the evaluation of video saliency is treated similar to image saliency. That is individual prediction maps are evaluated using image saliency evaluation scores. The scores are then averaged over all maps. For video saliency evaluation, we follow the steps of~\cite{wang2018cvpr} and use the same scores as provided in~\url{https://mmcheng.net/videosal/}, which are currently widely adopted scores. 

The scores include Normalized Scanpath Saliency (NSS), Similarity Metric (SIM), Linear Correlation Coefficient (CC), Area Under the ROC Curve (AUC-Judd)\cite{Judd2012}, and shuffled AUC (sAUC)~\cite{Zhang2008}. We sort the values based on the NSS metric as recomended by~\cite{salMetrics_Bylinskii}. A breif account of each score is provided for the sake of completeness.

The NSS is designed to evaluate a saliency map over fixation locations. Given a saliency map $S$ and a binary fixation map $F$, NSS is defined as,
\begin{align}
NSS = & \frac{1}{N} \sum_{i=1}^{N} \hat{S}(i) \times F(i), \\
\mbox{where} \quad  & N = \sum_i F(i), \quad \hat{S} = \frac{S - \mu}{\sigma}, \nonumber
\end{align}
\noindent where $\mu$ and $\sigma$ are the mean and standard  deviation of the saliency map.

The linear correlation coefficient (CC) measures the correlation between the saliency map $S$ and a smoothed fixation map $G$. It is defined as $CC = cov(S, G) / \sigma_S \times \sigma_G$, where $cov(S,G)$ is the covariance, $\sigma_S$ and $\sigma_G$ are the standard deviations of the saliency map and smoothed fixation map, respectively.
The similarity metric (SIM) measures the similarity between two distributions. Treating $S$ and $G$ as probability distributions,~\ie, ($\sum_i S(i) = 1$ and $\sum_i G(i)=1$), the SIM is computed as $SIM = \sum_i \min(S(i), G(i)$). 

The area under the curve (AUC) treats the estimated saliency map as a classifier output. It computes the area under the ROC curve obtained by varying a threshold and obtaining the true positive rate and false positive rate. The AUC-Judd samples the true positives from all the saliency map values above a threshold at fixated pixels and measures the false positive rate as the total saliency map values above a threshold at not-fixated pixels.
The sAUC samples the negatives from fixated locations of other images, other video frames in our case. This sampling scheme penalizes the center-bias phenomenon~\cite{Zhang2008}.

\subsection{Compared Baseline Models}

To understand the contribution of each modality, we propose two single modality models. The baseline models are using the same encoder architecture in a single stream, while the decoder is equivalent to the decoder of the multi-modal saliency model. In what follows, these baseline models will be referred to as `audio' model and `video' model. The two models are trained with the same setup as our proposed audio-visual model from the same starting point and weight initialization. Thus, we can assess and evaluate the contribution of the audio cue on a fair basis.

\subsection{Compared Video Saliency Models}
\label{sec:cmpmodel}

In order to gain an understanding about the performance of the proposed model in comparison to other video saliency models, we conduct a comprehensive evaluation. Previous works have demonstrated that video saliency models learning from spatio-temporal data outperform image saliency models in video saliency prediction~\cite{Jiang2018ECCV,wang2018cvpr}. We, thus, focus on video saliency models. Table~\ref{tab:cmodel} summarizes the information about the compared models.

The compared models consist of traditional video saliency models that rely on some image features or generic image statistics as well as state-of-the-art deep models. We could not compare our model with~\cite{CoutrotModel2016,multiECCV2018} since they do not have their code publicly available. Furthermore, they are task-specific and work only on sequences of speakers. 

Among deep models, we compare with two state-of-the-art works: DeepVS~\cite{Jiang2018ECCV} and ACLNet~\cite{wang2018cvpr} that have public implementations and pre-trained models available. In addition, for ACLNet, we fine tuned the model using AVE training data for five epochs starting from the pre-trained model with a learning rate of $1e^{-5}$ (DeepVS does not provide code for training). The fine-tuned model will be referred to as ACLNet*. 

Among traditional saliency models, we compared to SEO~\cite{Seo2009}, FEVS~\cite{TavakoliModel2013}, UHF-D~\cite{Wang_2017_CVPR_Workshops}, and AWS-D~\cite{AWSD2018}. Except UHF-D, which learns a set of filters by employing unsupervised hierarchical independent subspace analysis from a random video sequence from YouTube, the rest of the models are training-free. The models are developed for videos, though they may have image-only versions. The input to the models are the video frames of the same size as our deep model, \ie~$256\times320$. The temporal length of frames is equal to the models' expected number of frames. A model may resize the inputs internally when required.

\begin{table}[t]
\renewcommand{\arraystretch}{1.3}
 \caption{The list of the compared models, the source of the training data, the data used for extra fine-tuning (Ext. fin.), the type of the model (deep vs.~traditional) and use of center prior (`CB'). The fine tuning was only applied to deep models that provide the training scripts. The fine-tuned models are indicated by `*'. 'PO' stands for the use of center prior at prediction output and 'ML' indicates the use of prior in middle layers.}
 \label{tab:cmodel}
 \centering
 \small
 \begin{tabular}{l c c c c}
 \hline
 \textbf{Model} & \textbf{Train} & \textbf{Ext. fin.}  & \textbf{Deep} & \textbf{CB}\\
 \hline
 Audio (ours)& AVE & --- & \ding{51} & \ding{55}\\
 Video (ours) & AVE & --- & \ding{51} & \ding{55}\\
 Audio-Visual (ours) & AVE & --- & \ding{51}& \ding{55}\\
 \hline
 FEVS~\cite{TavakoliModel2013} & --- & --- & \ding{55} & PO\\
 AWS-D~\cite{AWSD2018} & --- & --- & \ding{55}  & \ding{55} \\
 SEO~\cite{Seo2009} & --- & --- & \ding{55} &  \ding{55}\\
 UHF-D~\cite{Wang_2017_CVPR_Workshops} & YouTube & --- & \ding{55} & PO\\
 \hline
 ACLNet~\cite{wang2018cvpr} & DHF1K & --- & \ding{51} &\ding{55}\\
 ACLNet*~\cite{wang2018cvpr}  & DHF1K & AVE & \ding{51} & \ding{55}\\
 DeepVS~\cite{Jiang2018ECCV} & LEDOV & --- & \ding{51} & ML\\
 \hline
 
 \end{tabular}
 \end{table}

\section{Results}

\subsection{Modal Integration Strategies}

How should we integrate the features obtained from two modalities?
It is expected that concatenating the features and learning the best combination with the decoder will result in the best performance.
To asses different strategies, we experiment with concatenation, element-wise multiplication, and element-wise summation of the features. The results are summarized in Table~\ref{tab:modalinteg}. As depicted, concatenation performs the best. In the rest of the paper, whenever the integration type is not explicitly mentioned, we are using concatenation for the audio-visual model.

\begin{table}[t]
\caption{Audio-visual model and modality integration strategies: performance of the proposed audio-visual model using different mechanisms for combining audio and video features.}
 \label{tab:modalinteg}
 \renewcommand{\arraystretch}{1.3}
 \renewcommand{\tabcolsep}{.9mm}  
 \centering
 \footnotesize
 \begin{tabular}{l c c c c c}
 \hline
  \textbf{Integ. type} & \textbf{NSS} $\uparrow$ & \textbf{AUC Judd} $\uparrow$  & \textbf{sAUC} $\uparrow$   & \textbf{CC} $\uparrow$   & \textbf{SIM} $\uparrow$ \\
 \hline
 Concat. & \textbf{2.45} & \textbf{0.8818} & \textbf{0.7268} & \textbf{0.5457} & \textbf{0.4493} \\
 Elm. Mul. & 2.36 & 0.8800 & 0.7265 & 0.5269 & 0.4309 \\
 Sum & 2.26 & 0.8729 & 0.7174 & 0.5077 & 0.42145 \\
 \hline
 \end{tabular}
 \end{table}

\subsection{Contribution of Modalities}

\textit{How much each modality and their combination contribute to predicting dynamic saliency?} To answer, we look into the contribution of individual modalities and their combination. To this end, we use the proposed baselines against our audio-visual model and report the results on each category and overall. We also include `MEP' and `Human Infinite' as the lower and upper-bounds of the data for better understanding. 

\begin{figure}
    \centering
    \includegraphics[scale=0.28]{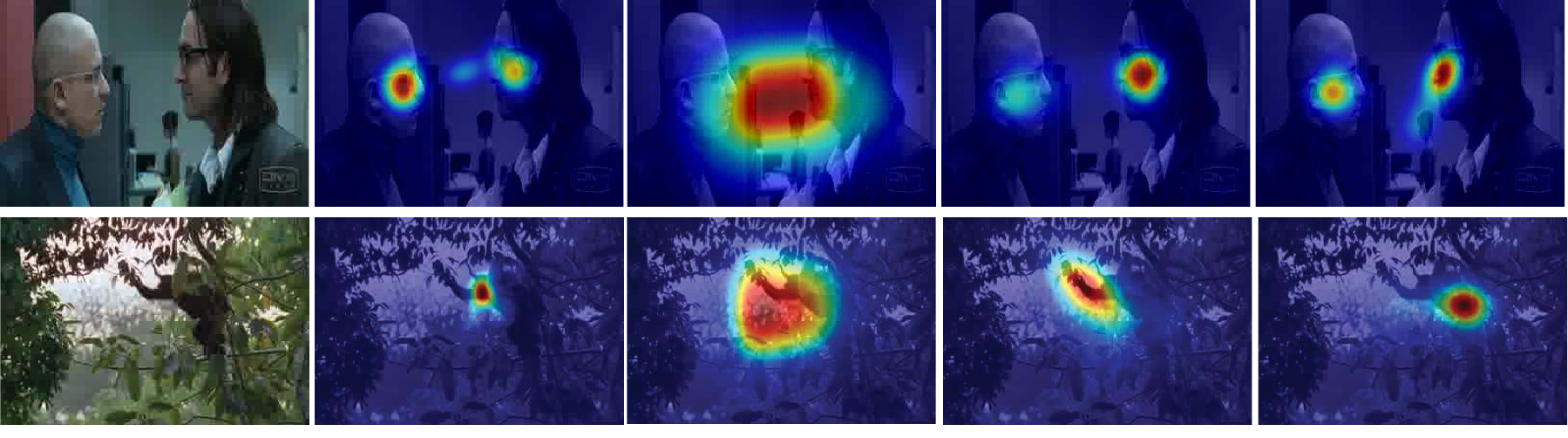}
    \caption{Contribution of each modality in model. From left to right: image, ground truth, audio model, video model, multimodal model. The audio model is focused on the center. The video model finds the correct spatial regions, while the multimodal model has a better attention distribution.}
    \label{fig:models}
\end{figure}

The results are summarized in Table~\ref{tab:percat}. The combination of both modalities results in the best saliency prediction scores. An ANOVA test on scores with each modality as factors indicates significant difference ($p < 0.05$) between all the models on all the scores. This shows that the combination of audio and visual cues as features indeed improves the saliency prediction significantly. Example maps are visualized in Fig.~\ref{fig:models}

\begin{table}[t]
\caption{Multimodal model and contribution of each modal with respect to video categories and overall. Mean eye position (MEP) is obtained from all videos in the training set of AVE and assessed against the videos in each category in the test set. Human Infinite is obtained by assessing half of the subjects against the other half in test split. MEP is a strong baseline and human infinite defines the expected upper bound of the test data. The best scores are shown in bold letters.}
 \label{tab:percat}
\renewcommand{\arraystretch}{1.3}
  \renewcommand{\tabcolsep}{.9mm}  

 \footnotesize
 \centering
 \begin{tabular}{c l c c c c c}
 \hline
 \textbf{Cat.}&\textbf{Model Name} & \textbf{NSS} $\uparrow$ & \textbf{AUC Judd} $\uparrow$   & \textbf{sAUC} $\uparrow$   & \textbf{CC} $\uparrow$   & \textbf{SIM} $\uparrow$ \\
 \hline
\multirow{4}{*}{\rotatebox{90}{\textbf{Nature}}} & Human Infinite & \textcolor{red}{3.31} & \textcolor{red}{0.8806} & 
\textcolor{red}{0.7724} & 
\textcolor{red}{0.6961} & 
\textcolor{red}{0.5603} \\
& Audio-Visual & \textbf{2.27} & \textbf{0.8773} & \textbf{0.7233} & 
\textbf{0.5392} & 
\textbf{0.4504} \\
& Video  & 2.04 & 0.8762 & 0.7191 & 0.5066 & 0.4073 \\
& MEP & \textcolor{blue}{1.76} & 
\textcolor{blue}{0.8696} & 
\textcolor{blue}{0.6864} & 
\textcolor{blue}{0.4714} & 
\textcolor{blue}{0.3682} \\
& Audio & 1.72 & 0.8684 & 0.6888 &  0.4637 &  0.3678 \\
 \hline
 \multirow{4}{*}{\rotatebox{90}{\textbf{Soc. Ev.}}} & Human Infinite & 
 \textcolor{red}{3.65} & 
 \textcolor{red}{0.8765} & 
 \textcolor{red}{0.7760} & 
 \textcolor{red}{0.6971} & 
 \textcolor{red}{0.5485} \\
& Audio-Visual & \textbf{2.65} & \textbf{0.8853} & 0.7264 & \textbf{0.5453} & \textbf{0.4420}\\
& Video & 2.45 & 0.8824 & \textbf{0.7275} & 0.5136 & 0.4080\\
& MEP &  \textcolor{blue}{1.35} & 
\textcolor{blue}{0.8196} & \textcolor{blue}{0.6337} & 
\textcolor{blue}{0.3147} & 
\textcolor{blue}{0.2744} \\
& Audio & 1.32 & 0.8180 & 0.6322    & 0.3123 &  0.2756 \\
 \hline
 \multirow{4}{*}{\rotatebox{90}{\textbf{Misc.}}} & Human Infinite & 
 \textcolor{red}{3.34} & 
 \textcolor{red}{0.8682} & 
 \textcolor{red}{0.7716} & 
 \textcolor{red}{0.6543} & 
 \textcolor{red}{0.5256} \\
& Audio-Visual & \textbf{2.39} & \textbf{0.8812} & 0.7360 & \textbf{0.5495} & \textbf{0.4548} \\
& Video & 2.25 & 0.8774 & \textbf{0.7373} & 0.5321 & 0.4335 \\
& MEP & \textcolor{blue}{1.73} & 
\textcolor{blue}{0.8446} & 
\textcolor{blue}{0.6754} & 
\textcolor{blue}{0.4378} & 
\textcolor{blue}{0.3422} \\
& Audio & 1.60 & 0.8400 & 0.6714 &   0.4114 &   0.3368 \\
 \hline
  \multirow{4}{*}{\rotatebox{90}{\textbf{Overall}}} &Human Infinite & \textcolor{red}{3.83} & \textcolor{red}{0.8852} &
 \textcolor{red}{0.7877} & 
 \textcolor{red}{0.7140} &
 \textcolor{red}{0.574 } \\
 &Audio-Visual  & \textbf{2.45} & \textbf{0.8818} & \textbf{0.7268} & \textbf{0.5457} & \textbf{0.4493} \\
&Video & \underline{2.26} & \underline{0.8793} & \underline{0.7259} & \underline{0.5178} & \underline{0.4158}\\ 
 &MEP  & 
 \textcolor{blue}{1.59} & 
 \textcolor{blue}{0.8443} & 
 \textcolor{blue}{0.6623} & 
 \textcolor{blue}{0.4029} & 
 \textcolor{blue}{0.3263} \\
&Audio  & 1.54 & 0.8419 & 0.6631 & 0.3929 & 0.3246\\
\hline
 \end{tabular}
 \end{table}

 \textit{Why audio model performs worst than MEP?}
The Auido-Visual model and Video model are performing significantly better than MEP. Nevertheless, the audio signal falls short of MEP in terms of scores. Furthermore, as depicted in Fig.~\ref{fig:models}, the visual inspection of audio model shows that its prediction maps are mostly focused on the center of frames, despite its statistical difference with MEP (ANOVA, $p < 0.05$).

The reason lies on the input dimension and spatial resolution of the features obtained from the encoder. The audio stream accepts the inputs of shape $16\times3\times64\times64$ and produces a feature vector with spatial size of $1024\times1\times1$, which is replicated to the spatial size of the visual features. The configuration is motivated by the fact that we are not expecting audio features having a spatial interpretation by themselves and they rather carry a temporal aspect to the presence of a salient element that sounds. In such audio model, the center focus is, thus, reasonable given the fact that the visual spatial information channel is absent and consequently the precise spatial localization of the sound source in the spatial domain is difficult if not impossible.

\subsection{Center-bias}

\textit{To bias or not to bias?}
There has been a long-standing debate weather one should exploit the center-bias in the fixation data or not and the matter is still not settled.

Regardless, to get an idea about the performance of the model with center-bias, we provide the overall performance of the proposed model and baselines with the MEP included. For this purpose, we learn the MEP from all the training sequences and add it to the model similar to the DeepGazeII~\cite{deepgaze2} approach for images. 

Table~\ref{tab:resmep} summarizes the results of our models with and without MEP. We find that adding center-bias to all models
improves their performance over all scores, except sAUC, which penalizes the center-bias in models. The Audio-Visual+MEP and Audio-Visual model are the best two performing models. Nonetheless, for the case of the Audio model, we do not observe any improvement mostly due to the central focus of the audio model.
Overall, the Audio-Visual model that penalizes center-bias has the best sAUC score. This, along with other competitive scores, indicates that our Audio-Visual model is superior in capturing attention without the incorporation of any center-bias.

 \begin{table}[t]
  \caption{Performance of our models with/without center bias (MEP).
 }
 \label{tab:resmep}
 \renewcommand{\arraystretch}{1.3}
 \renewcommand{\tabcolsep}{.9mm}  
 \centering
  \begin{tabular}{l c c c c c}
 \hline
 \textbf{Model Name} & \textbf{NSS} $\uparrow$ & \textbf{AUC Judd} $\uparrow$  & \textbf{sAUC} $\uparrow$   & \textbf{CC} $\uparrow$   & \textbf{SIM} $\uparrow$ \\
 \hline
 Human Infinite & \textcolor{red}{3.83} & \textcolor{red}{0.8852} &
 \textcolor{red}{0.7877} & 
 \textcolor{red}{0.7140} &
 \textcolor{red}{0.574 } \\
Audio-Visual+MEP & \textbf{2.47} & \textbf{0.8842} & {0.7224} & \textbf{0.5505} & \textbf{0.4566} \\
Audio-Visual  & \underline{2.45} & {0.8818} & \textbf{0.7268} & \underline{0.5457} & \underline{0.4493} \\
Video+MEP  & {2.31} & \underline{0.8824} & \underline{0.7227} & {0.5283} & {0.4294}\\
Video  & {2.26} & {0.8793} & {0.7259} & {0.5178} & {0.4158}\\
 MEP  & 
 \textcolor{blue}{1.59} & 
 \textcolor{blue}{0.8443} & 
 \textcolor{blue}{0.6623} & 
 \textcolor{blue}{0.4029} & 
 \textcolor{blue}{0.3263} \\
Audio   & 1.54 & 0.8419 & 0.6631 & 0.3929 & 0.3246\\
Audio+MEP   & 1.54 & 0.8419 & 0.6586 & 0.3898 & 0.3168\\
 \hline
 \end{tabular}

 \end{table}
 
\subsection{Audio vs. MEP}

\textit{How effective is audio and does it differ from MEP (center-bias)?} The audio signal is effective and distinguishable from center-bias. The key difference between center-bias and audio model, despite central focus of audio due to its implementation scheme, is that the predictions of audio model is driven by stimulus, while the MEP is not stimulus driven and is fixed to represent the fixation bias in the data. Thus, it is expected that the combination of audio features with visual features as shown in Audio-Visual model outperform the other models with and without MEP.

Table~\ref{tab:resmep} provides further numerical evidence in support of the effectiveness of audio. Comparing Audio-Visual+MEP and Video+MEP, we observe Audio-Visual+MEP significantly outperforms the Video+MEP (ANOVA $p < 0.05$). Thus, the audio cue combined with visual cue improves the saliency prediction significantly in presence of center-bias. Furthermoe, The Video+MEP falls short of Audio-Visual model on four scores, which indicates that audio cue is powerful enough to empower the Audio-Visual model outperform video+MEP.

\subsection{Behaviour of the Models over Time}

\begin{figure}[t]
    \centering
    \includegraphics[scale=0.56]{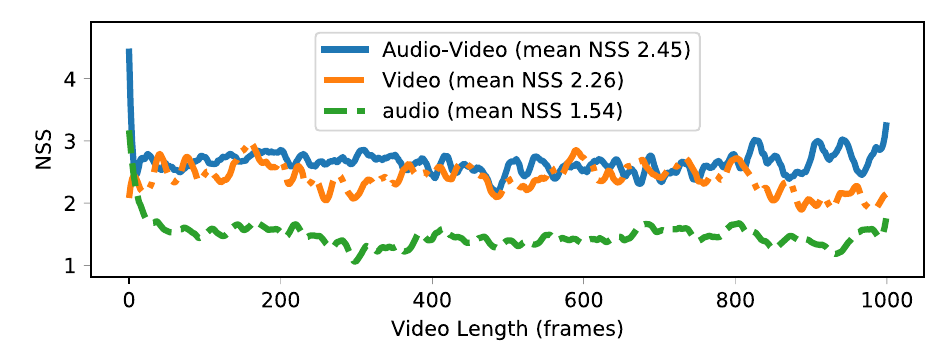}
    \includegraphics[scale=0.56]{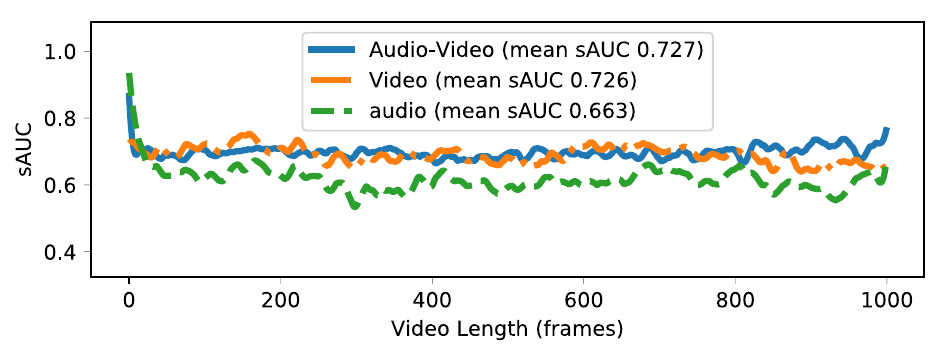}
    \caption{Mean NSS and sAUC of modalities over time. The values are smoothed for the visualization purposes. The overall performance, prior to re-sampling, of each modality is also reported.}
    \label{fig:meanbehave}
\end{figure}

\textit{What is the behavior of models over video duration?}
To understand the performance over video duration, we plot the scores of each frame over the video length (ordered frames) for mean performance and case-by-case (best and worst sequences per category). The average performance behavior over time is obtained by normalizing the length of videos (each video has different number of frames and duration). We re-sample the score of frames over video length to a fixed number, here 1000. Then, we compute the mean of scores of all the videos for each frame as in Fig.~\ref{fig:meanbehave}. 

Looking at the overall mean performance of the models over time, we learn that the Audio-Visual model has an edge over Audio and Video models most of the time. Comparing the NSS score of the Audio-Visual predictions with the video model over all the predicted frames, 53.54\% of the time the audio-visual model outperforms the video model (\ie~NSS score is higher). There are, however, cases where the video model performs slightly better than the audio-visual model.

\begin{figure*}[t]
    \centering
    \includegraphics[scale=0.38]{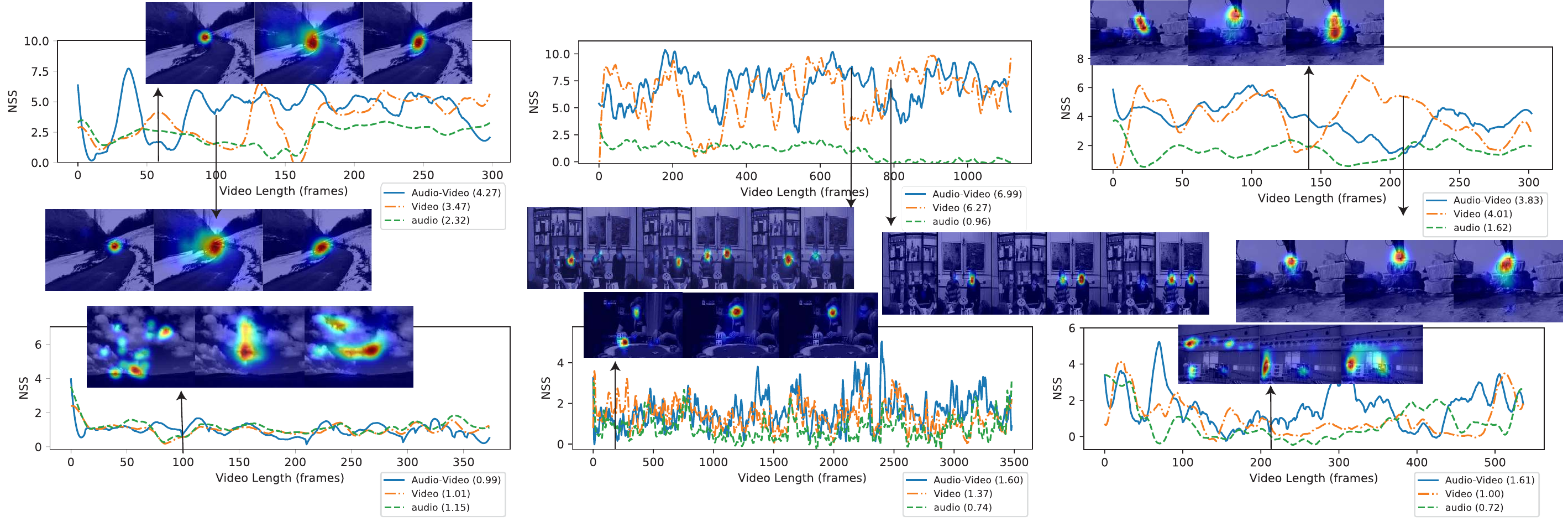}
    \caption{Comparing predictions over time between the proposed audio-video model and video model. From left to right: nature, social events, and miscellaneous categories. Top row sequences with best NSS score and  bottom row sequences with worst NSS score. Sample frames depict ground-truth, video model, and audio-video model maps overlaid on the frame (from left to right). }
    \label{fig:coherence}
\end{figure*}

In order to dig deeper into the cases where one model is better, we looked into the performance of the best and worst sequences for each category using the NSS score. The results are summarized in Fig.~\ref{fig:coherence}.
Investigating video frames and predictions suggest that the audio-visual model is more sensitive to active sound sources, whereas the video model is more responsive to visual cues. 
In Fig.~\ref{fig:coherence}, the middle column frames show examples in which the audio-visual model localizes the speaking face well, while the video model captures many faces. Thus, we look further into the role of sound source later.
To sum up, preferring one cue over the other for computing saliency leads to nuance performance variations over the video length, though audio-visual model has the overall advantage. 

\textit{Why all models perform similarly at the beginning of sequences?}
All the models have roughly equal performance at the beginning of the video sequences. To find out the reason, we sampled the first 10 frames from each video (290 frames) and checked the predictions and ground-truth visually, and in terms of scores. The analysis indicated that the ground truth data is mostly focused on the center of frames (78.97\% \ie~229 out of 290 frames) and models predict such cases easily (mean NSS score $> 2.2$).

\subsection{Impact of Sound Source Location}
\begin{figure}[h]
    \centering
    \includegraphics[scale=0.5]{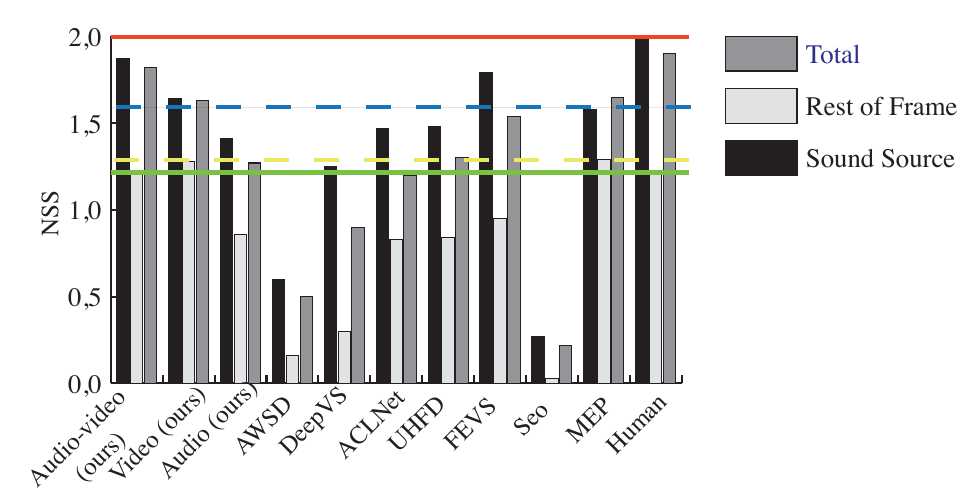}
    \caption{Contextual evaluation of the saliency models in the case of visible sound source. The red and green lines indicate human infinite performance over (1) the location of sound source and (2) other parts of the frame except sound source location, respectively. Similarly, the blue and yellow lines correspond to MEP performance on the cases (1) and (2), respectively. 
    }
    \label{fig:contextual}
\end{figure}

\textit{How well Audio-Visual model compares to video-based methods in localizing the visually salient sound source?} It is naturally expected that audio-visual saliency models perform well on cases where the video frame contains a salient visible sound source. We also noticed that our Audio-Visual model performed particularly well on cases where the video frame contains a visible sound source. To further assess this observation, we performed a contextual evaluation on the predicted saliency maps. To this end, we randomly sampled frames from each test video and manually annotated the location of the sound source if present. Only frames where a sound source was found were included in the subsequent analysis. Following the contextual evaluation scheme of~\cite{TavakoliABL17}, where a score is computed over regions of interest, we computed contextual NSS scores for all the models, including other video models. That is, we compute the NSS score over (1) the location of the sound source, (2) other parts of the frame except the sound source location, and (3) the entire test frame. Results are reported in Fig.~\ref{fig:contextual}. As depicted, the audio-visual model has the best contextual performance, attending to the sound source consistent with human attention. This finding is aligned with the findings of human studies where a synchronous audio-visual stimulus influences and grabs gaze~\cite{Burg2008}.

\subsection{Comparison to Video Saliency Models}
\begin{figure*}[h]
    \centering
    \includegraphics[scale=0.36]{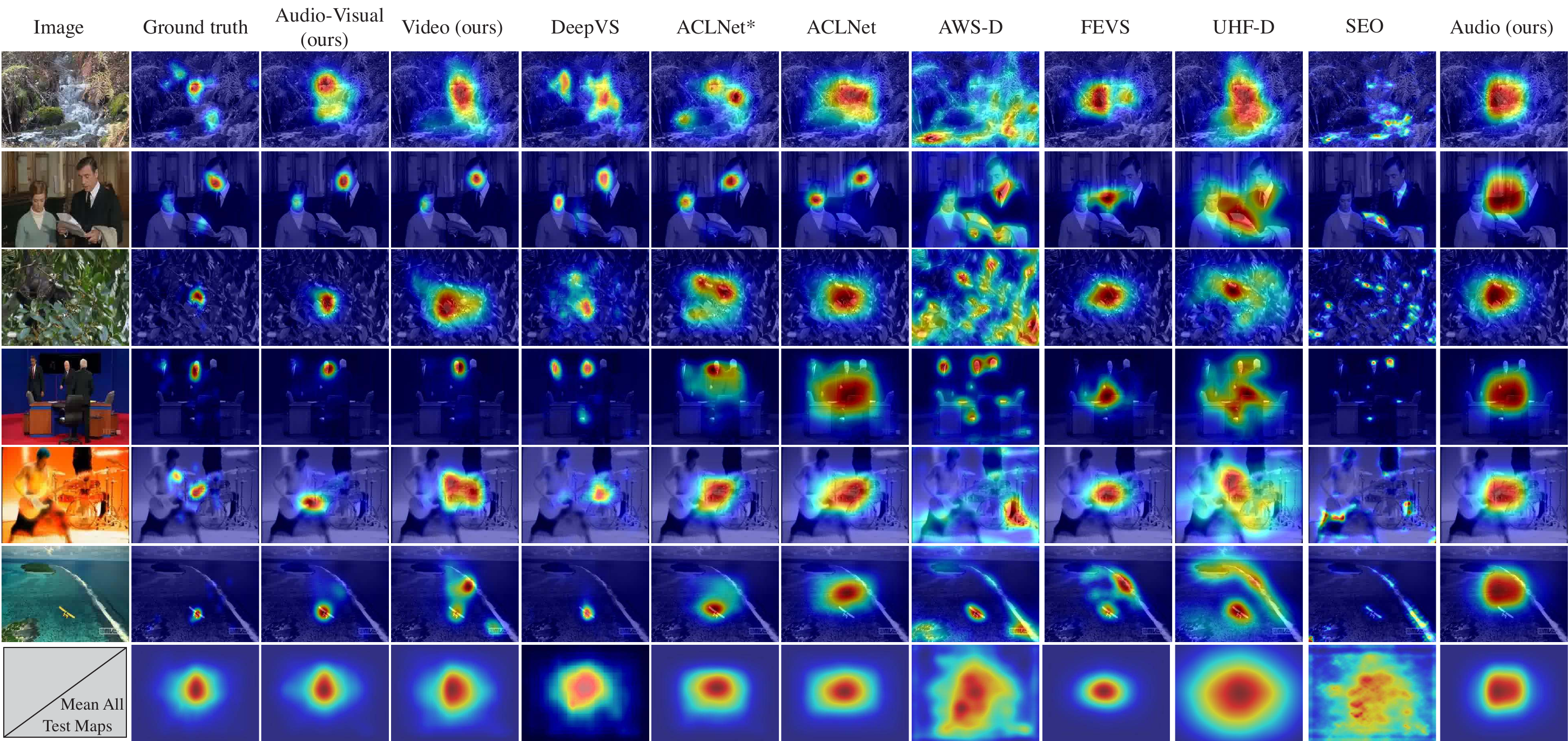}
    \caption{Example saliency maps from the tested models. The last row contains the mean saliency map of each model over the test frames.}
    \label{fig:salmaps}
\end{figure*}

\textit{How well the Audio-Visual model compares to existing video models?}
We compare the proposed Audio-Visual model to video models overall and over each category. Table~\ref{tab:results} summarizes the performance of all models. Our comparison shows that the proposed Audio-Visual model outperforms all other models. Our proposed Video model also outperforms the other video models. Fine-tuning ACLNet improved its performance, albeit small. This indicates that possibly deep models are already generalized enough to handle any unseen visual data, which alleviates our concerns regarding the lack of access to DeepVS training code for fine-tuning it on AVE. Deep video models perform better than the tested traditional saliency models. Most of the traditional models fall short in comparison to MEP on majority of the scores, except FEVS. Fig.~\ref{fig:salmaps} depicts some randomly chosen predictions to facilitate better understanding of models' output. As shown, the proposed audio-visual model produces maps that are more similar to the ground truth human density maps. We also visualize the mean maps over the test frames.

 \begin{table}[t]
 \renewcommand{\arraystretch}{1.3}
 \renewcommand{\tabcolsep}{.9mm}  
 \caption{Performance of various models on AVE test set, sorted by NSS. $(\uparrow)$ indicates that higher value is better. Mean Eye Position (MEP) from training center indicates a strong baseline and Human Infinite defines the upper performance bound. The best score is in bold, and the second best is underlined.}
 \label{tab:results}
 \centering
 \footnotesize
 \begin{tabular}{l c c c c c}
 \hline
 \textbf{Model Name} & \textbf{NSS} $\uparrow$ & \textbf{AUC Judd} $\uparrow$  & \textbf{sAUC} $\uparrow$   & \textbf{CC} $\uparrow$   & \textbf{SIM} $\uparrow$ \\
 \hline
 Human Infinite & \textcolor{red}{3.83} & \textcolor{red}{0.8852} &
 \textcolor{red}{0.7877} & 
 \textcolor{red}{0.7140} &
 \textcolor{red}{0.574 } \\
 \hline
Audio+Visual (ours) & \textbf{2.45} & \textbf{0.8818} & \textbf{0.7268} & \textbf{0.5457} & \textbf{0.4493} \\
Video (ours) & \underline{2.26} & \underline{0.8793} & \underline{0.7259} & \underline{0.5178} & \underline{0.4158}\\
 DeepVS & 2.05 & 0.8603 & 0.6996 & 0.4666 & 0.3733\\
 ACLNet* & 1.98 & 0.8698 & 0.7003 & 0.4755 & 0.3793\\
 ACLNet & 1.88 & 0.8670 & 0.6960 & 0.4625 & 0.3489\\
  FEVS & 1.71 & 0.8532 & 0.6777 & 0.4217 & 0.3629 \\
  \hline
 MEP  & 
 \textcolor{blue}{1.59} & 
 \textcolor{blue}{0.8443} & 
 \textcolor{blue}{0.6623} & 
 \textcolor{blue}{0.4029} & 
 \textcolor{blue}{0.3263} \\
  \hline
  UHF-D & 1.56 & 0.8401 & 0.6945 & 0.38104 & 0.2752 \\
 AWS-D & 0.99 & 0.7309 & 0.6537 & 0.2317 & 0.2248 \\
 SEO & 0.55 & 0.6899 & 0.5512 & 0.1276 & 0.1946 \\
 \hline
 \end{tabular}
 \end{table}

The scores for each category are provided in Table~\ref{tab:modelpercat}.
The results indicate that the proposed Audio-Visual model and our baseline Video model outperform other dynamic saliency models over individual categories. We looked also into the percentage of frames that the Audio-Visual model outperforms other models. Results are reported in Table~\ref{tab:nssimprove}, and indicate significant improvements.

  \begin{table}[t]
  \renewcommand{\arraystretch}{1.3}
 \caption{Performance of various models on AVE test set cateories, sorted by NSS. $(\uparrow)$ indicates that higher value is better. Mean Eye Position (MEP) from training center indicates a strong baseline and Human Infinite defines the upper performance bound. The best score is in bold, and the second best is underlined.
 }
 \label{tab:modelpercat}
  \renewcommand{\tabcolsep}{.9mm}  

 \footnotesize
 \centering
 \begin{tabular}{c l c c c c c}
 \hline
 \textbf{Cat.}&\textbf{Model Name} & \textbf{NSS} $\uparrow$ & \textbf{AUC Judd} $\uparrow$   & \textbf{sAUC} $\uparrow$   & \textbf{CC} $\uparrow$   & \textbf{SIM} $\uparrow$ \\
 \hline
\multirow{12}{*}{\rotatebox{90}{\textbf{Nature}}} & Human Infinite & \textcolor{red}{3.31} & \textcolor{red}{0.8806} & 
\textcolor{red}{0.7724} & 
\textcolor{red}{0.6961} & 
\textcolor{red}{0.5603} \\
& Audio-Visual (ours)& \textbf{2.27} & \textbf{0.8773} & \textbf{0.7233} & 
\textbf{0.5392} & 
\textbf{0.4504} \\
& Video (ours) & \underline{2.04} & 
\underline{0.8762} & 
\underline{0.7191} & 
\underline{0.5066} & 
\underline{0.4073} \\
& DeepVS & 1.89 & 0.8569 & 0.6959 & 0.4621 & 0.3751 \\
& ACLNet* & 2.03 & 0.8841 & 0.7232 & 0.5179 & 0.4010 \\
& ACLNet & 1.96 & 0.8811 & 0.7164 & 0.5116 & 0.3766 \\
& FEVS & 1.85 & 0.8719 & 0.7019 & 0.4816 & 0.3991 \\
& MEP & \textcolor{blue}{1.76} & 
\textcolor{blue}{0.8696} & 
\textcolor{blue}{0.6864} & 
\textcolor{blue}{0.4714} & 
\textcolor{blue}{0.3682} \\
& UHF-D & 1.64 & 0.8456 & 0.7103 & 0.4153 & 0.2943 \\
& AWS-D & 0.74 & 0.6799 & 0.6288 & 0.1723 & 0.2092 \\
& SEO & 0.45 & 0.6578 & 0.5434 & 0.1021 & 0.1881 \\
 \hline
 \multirow{12}{*}{\rotatebox{90}{\textbf{Soc. Ev.}}} & Human Infinite & 
 \textcolor{red}{3.65} & 
 \textcolor{red}{0.8765} & 
 \textcolor{red}{0.7760} & 
 \textcolor{red}{0.6971} & 
 \textcolor{red}{0.5485} \\
& Audio-Visual (ours) & \textbf{2.65} & \textbf{0.8853} & \underline{0.7264} & \textbf{0.5453} & \textbf{0.4420}\\
& Video  (ours) & \underline{2.45} & \underline{0.8824} & \textbf{0.7275} & \underline{0.5136} & \underline{0.4080}\\
& DeepVS & 2.26 & 0.8671 & 0.7008 & 0.4775 & 0.3723 \\
& ACLNet* & 2.02 & 0.8692 & 0.6834 & 0.4488 & 0.3594 \\
& ACLNet & 1.91 & 0.8659 & 0.6837 & 0.4324 & 0.3251 \\
& FEVS & 1.56 & 0.8393 & 0.6546 & 0.3540 & 0.3210 \\
& UHF-D & 1.46 & 0.8301 & 0.6675 & 0.3364 & 0.2502 \\
& MEP &  \textcolor{blue}{1.35} & 
\textcolor{blue}{0.8196} & \textcolor{blue}{0.6337} & 
\textcolor{blue}{0.3147} & 
\textcolor{blue}{0.2744} \\
& AWS-D & 1.13 & 0.7658 & 0.6575 & 0.2656 & 0.2299 \\
& SEO & 0.60 & 0.7137 & 0.5488 & 0.1404 & 0.1936 \\
 \hline
 \multirow{12}{*}{\rotatebox{90}{\textbf{Misc.}}} & Human Infinite & 
 \textcolor{red}{3.34} & 
 \textcolor{red}{0.8682} & 
 \textcolor{red}{0.7716} & 
 \textcolor{red}{0.6543} & 
 \textcolor{red}{0.5256} \\
& Audio-Visual (ours) & \textbf{2.39} & \textbf{0.8812} & \underline{0.7360} & \textbf{0.5495} & \textbf{0.4548} \\
& Video (ours) & \underline{2.25} & \underline{0.8774} & \textbf{0.7373} & \underline{0.5321} & \underline{0.4335} \\
& DeepVS & 1.98 & 0.8555 & 0.7027 & 0.4574 & 0.3723 \\
& ACLNet* & 1.84 & 0.8517 & 0.6831 & 0.4562 & 0.3778 \\
& ACLNet & 1.74 & 0.8497 & 0.6858 & 0.4391 & 0.3450 \\
& FEVS & 1.74 & 0.8473 & 0.6773 & 0.4354 & 0.3728 \\
& MEP & \textcolor{blue}{1.73} & 
\textcolor{blue}{0.8446} & 
\textcolor{blue}{0.6754} & 
\textcolor{blue}{0.4378} & 
\textcolor{blue}{0.3422} \\
& UHF-D & 1.59 & 0.8465 & 0.7108 & 0.3971 & 0.2843 \\
& AWS-D & 1.13 & 0.7658 & 0.6575 & 0.2656 & 0.2299 \\
& SEO & 0.61 & 0.6996 & 0.5647 & 0.1436 & 0.2045 \\
 \hline
 \end{tabular}
 \end{table}
 
\textit{Is the comparison between our video model and existing models fair?}
There is a natural difference in our architecture in comparison to other learning based approaches. That is, we are employing 3D ResNets pre-trained on a large video corpus, while other models such as ACLNet and DeepVS rely on image-based backbones. Interestingly, we are achieving a good performance with a simpler neural architecture and loss function. This result is indeed well-aligned with other works that are reporting integration of data from various sources can lead to better saliency prediction because of the superior feature representations that exist in such encoders~\cite{EMLNet2018}. Our architecture follows the feature integration theory and achieves better performance by exploiting a superior feature representation. This superiority is further reinforced with the audio features in the Audio-Visual model.

\begin{table}
\renewcommand{\arraystretch}{1.3}
    \caption{Percentage of frames with better NSS score for the Audio-Visual model in comparison to other models.}
    \label{tab:nssimprove}
    \centering
    \begin{tabular}{l|c}
    \hline
        \textbf{Compared Model} & \textbf{\% improved frame prediction}  \\
        \hline
        Video (ours) &	53.54 \\
        DeepVS & 60.46\\
        ACLNet*	& 67.18 \\
        FEVS &	68.05 \\
        UHFD &	69.31 \\
        ACLNet & 70.94 \\
        Audio (ours) & 73.13 \\
        AWSD &	82.04 \\
        SEO &	90.36 \\
        \hline
    \end{tabular}
\end{table}

\section{Discussion and Future Work}

In this paper, we presented a generic Audio-Visual deep model for dynamic saliency prediction and demonstrated the effectiveness of audio as a cue contributing to attention prediction using such model. The model is designed to be consistent with feature integration theory. The architecture of our model is simple and facilitates the study of cue (audio or video) contribution using two baseline models.

\textit{Audio cue is an effective factor in dynamic saliency prediction.} We did an extensive analysis that shows audio is a contributing factor that improves saliency prediction. The Audio-Visual model significantly performs better than the video baseline model on all scores overall and over each category. Further, analysis of the localization of sound source demonstrates that a model equipped with audio cue (Audio-Visual model) significantly improves over a video only capable model (Video model) in predicting the active sound source that is visually present. Our findings is well aligned with cognitive and behavioral studies, \eg~\cite{Burg2008,Coutrot2011erbsound}.

\textit{Richer feature representations enhance saliency prediction.} Analysis of our Video model demonstrates that rich feature representations, \eg~3D CNNs pre-trained on large scale video databases, are potentially better predictors of dynamic saliency. This is aligned with the research demonstrating exploiting features from multiple sources, \ie~scene recognition and image recognition, improves saliency prediction for images~\cite{EMLNet2018}. While the current trend in deep saliency models has been developing more sophisticated decoders for saliency prediction, it seems enriching feature representations is yet a potentially important aspect that needs further exploration.

\textit{More sophisticated Audio-Visual saliency models should be researched.} We kept our architecture simple to enhance analysis of audio cue in comparison with two simple baselines. Our model is very simple and consistent with FIT. There is still significant potential for trying more sophisticated decoders for Audio-Visual model. Nevertheless, that may require a large corpus of audio visual eye tracking database. It is note worthy that audio saliency models also require further investigation. There is significant potential for researching more biologically inspired audio models and their integration with the visual pathway.  

\textit{Large audio visual eye-tracking databases are helpful.} We circumvent the lack of large-scale audio visual data by pooling together data from various sources, pre-processing and categorizing them. The efforts on extending `AVE' corpus will furthermore facilitate learning end-to-end trainable deep models and facilitate more in-depth research on Audio-Visual attention. To extend the `AVE' corpus, one should consider stimulus diversity and potential use cases such as augmented and virtual reality.

\textit{We need to study the temporal aspects of human attention carefully.} Analysis of models over time showed that their behavior is very similar at the beginning of the videos. The root cause seemed to lie on the nature of the ground truth data, where fixations over 78\% of the first 10 frames of each video are concentrated at the center. We need to further investigate this phenomenon from a human behaviour aspect. Is this due to the lag in the attention system (subjects need some time to understand the story)? 

\textit{More diverse contextual annotation will help understanding the Audio-Visual attention better.}
Investigating the performance over the video length hinted to variations in favouring different input cues (audio and visual) in our audio-video model versus the video model. In-depth analysis of these fine variations is a future step, which requires controlled human experiments with extensive contextual annotation.

\textit{Better evaluation schemes are needed for video saliency.}
In this work, we used the current widely adopted method for video saliency benchmarks~\cite{wang2018cvpr}. It is an extension of image saliency techniques. Nevertheless, the nature of video saliency is different. It remains an open question weather one has to consider the sequence of predictions and how one should report the temporal behaviour of a model.

\section{Conclusion}

We introduced a deep end-to-end trainable audio-visual saliency model for predicting fixations over dynamic input. To train such model, we pooled together several existing databases under the umbrella of `AVE' corpus. We demonstrated that audio signal contributes significantly to dynamic saliency prediction. Audio-visual saliency models can be utilized in applications centered around dynamic attention with audio as their ubiquitous nature,~\eg~augmented reality, virtual and mixed reality, and human-robot interaction. 

We also discussed and hinted several directions for future studies and steps towards better understanding audio-visual attention. We also discussed potential steps for more advanced models.

\ifCLASSOPTIONcaptionsoff
  \newpage
\fi



\bibliographystyle{IEEEtran}
\bibliography{vidsal}
\end{document}